\documentclass[conference]{IEEEtran}
\IEEEoverridecommandlockouts
\usepackage{cite}
\usepackage{booktabs}
\usepackage{amsmath,amssymb,amsfonts}
\usepackage{algorithm}
\usepackage{algorithmic}
\usepackage{graphicx}
\usepackage{textcomp}

\usepackage{xcolor}
\usepackage{xspace} 
\usepackage{url}
\usepackage{hyperref}
\usepackage{array}
\usepackage{changepage}
\def\BibTeX{{\rm B\kern-.05em{\sc i\kern-.025em b}\kern-.08em
    T\kern-.1667em\lower.7ex\hbox{E}\kern-.125emX}}
\newcommand{\ie}{, i.e.,\xspace}

\newcommand{\eg}{, e.g.,\xspace}

\usepackage{cleveref}

\begin{document}

\title{Causal Knowledge Transfer for Multi-Agent Reinforcement Learning in Dynamic Environments}

\author{\IEEEauthorblockN{Kathrin Korte*, Christian Medeiros Adriano\textsuperscript{\textdagger}, Sona Ghahremani\textsuperscript{\textdagger}, Holger Giese\textsuperscript{\textdagger}}\IEEEauthorblockA{\textit{IT University of Copenhagen, Denmark}*~~~~~~\textit{Hasso Plattner Institute, University of Potsdam, Germany}\textsuperscript{\textdagger}\\Email:  kort@itu.dk*,~\{firstname.lastname\}@hpi.de\textsuperscript{\textdagger}}}

\maketitle
\begin{abstract}
[Context] Multi-agent reinforcement learning (MARL) has achieved notable success in environments where agents must learn coordinated behaviors. However, transferring knowledge across agents remains challenging in non-stationary environments with changing goals. [Problem] Traditional knowledge transfer methods in MARL struggle to generalize, and agents often require costly retraining to adapt. [Approach] This paper introduces a \emph{causal knowledge transfer framework} that enables RL agents to learn and share compact causal representations of paths within a non-stationary environment. As the environment changes (new obstacles), agents' collisions require adaptive recovery strategies. We model each collision as a \emph{causal intervention} instantiated as a sequence of \emph{recovery actions} (a macro) whose effect corresponds to a causal knowledge of how to circumvent the obstacle while increasing the chances of achieving the agent's goal (maximizing cumulative reward). This \textit{recovery action macro} is transferred online from a second agent and is applied in a zero-shot fashion, i.e., without retraining, just by querying a lookup model with local context information (collisions). [Results] Our findings reveal two key insights: (1) agents with heterogeneous goals were able to bridge about half of the gap between random exploration and a fully retrained policy when adapting to new environments, and (2) the impact of causal knowledge transfer depends on the interplay between environment complexity and agents' heterogeneous goals.
\end{abstract}

\section{Introduction }
\noindent\textbf{Context} 
A central mechanism underpinning intelligent behavior is the internal representations of the world that substitute for direct sensory input and support planning, prediction, and action based on learned regularities~\cite{adami2023elements}. However, in the process of adapting to an evolving world and changing goals~\cite{bloembergen2015evolutionary}, reinforcement learning (RL) agents~\cite{sutton2018reinforcement} can suffer from catastrophic forgetting~\cite{kumaran2016learning} or negative transfer~\cite{zhang2022survey}. As mitigation, RL agents can learn internal representations that encode a causal structure of the world, which is more stable (invariant to perturbations), hence enabling agents to anticipate action outcomes and reason counterfactually~\cite{deng2023causal}. 

\noindent\textbf{Problem} While multi-agent reinforcement learning (MARL) has enabled agents to learn effective behaviors in shared environments, robust knowledge transfer among agents remains an unsolved problem. Agents trained in one scenario often fail to generalize when faced with minor changes in their goals or the environment. 
This brittleness to change hinders MARL’s scalability\cite{yadav2023comprehensive} 
and real-world deployment\cite{zhou2024multiagent}. 
Despite the large body of work on knwoledge transfer methods 
 for MARL~\cite{zhao2023less,multi_agent_ride_sharing_systems_2021,multi_agent_transfer_rl_traffic_2021,multi_agent_common_knowledge_rl_2019}, current transfer methods, such as copying policies or value functions, tend to be brittle when applied to new contexts or altered dynamics. The reason is that they often lack the semantic structure needed for general reasoning and do not exploit deeper patterns across agents' experiences.
\noindent\textbf{State-of-the-Art} Most transfer learning methods in MARL are grounded in model-free RL paradigms, which learn action values or policies from direct interactions with the environment. While effective in narrow settings, these methods often lack explanatory structure and struggle with sample efficiency, especially in changing environments. Recent work in causal representation learning has shown that encoding cause-effect relationships can enhance generalization and robustness \cite{mutti2023provably}. Causal models offer a principled framework for distinguishing stable knowledge, e.g., effects of obstacles, from context-specific policies, thereby supporting more reliable transfer~\cite{adriano2024principled,AdrianoNSCausal}. Nonetheless, applying causal reasoning in MARL remains relatively unexplored. Few existing approaches treat individual agent experiences as data points in structured experiments from which causal effects can be estimated and transferred~\cite{briglia2024improving}.
%
%

\noindent\textbf{Approach} This paper presents a causal transfer learning framework for MARL in a discrete grid-world with dynamic configuration and partially overlapping goal structures. Agents independently explore and encounter disruptions (collisions) that affect task completion. These disruptions are treated as \emph{structural causal experiments}, enabling the discovery of \textit{recovery macros}—precomputed strategies that counteract specific context-dependent effects. 
Stored in a context-indexed lookup model, these macros enable zero-shot transfer: subsequent agents can apply them immediately in similar contexts, without additional training, facilitating rapid adaptation to novel environments and heterogeneous goals. The approach also supports decentralized coordination in MARL, allowing agents to share discoveries and improve collectively without centralized planning. 
 We investigate our approach with four key research questions:
\textbf{RQ1}\textit{ -- Does environment complexity impact the efficiency of causal-based knowledge transfer?}
\textbf{RQ2}\textit{ -- Does heterogeneity of goals impact the efficiency of causal-based knowledge transfer?} 
\textbf{RQ3}\textit{ -- Is the difference in causal-based knowledge transfer sensitive to environment complexity?} 
\textbf{RQ4} \textit{ -- Is the difference in causal-based knowledge transfer sensitive to goal heterogeneity?} 
By leveraging underlying causal structures rather than policy/value-based transfer, we aim to create a foundation for sample-efficient transfer learning in dynamic, heterogeneous multi-agent environments.

\noindent\textbf{Tour}~\Cref{sec:preliminaries} details the preliminaries and \Cref{sec:state-of-the-art} presents the state-of-the-art. \Cref{sec:approach} discusses the causal knowledge transfer approach. \Cref{sec:evaluation} shows the evaluation and empirical results and \Cref{sec:discussion-conclusion} concludes the paper.

\section{Preliminaries}\label{sec:preliminaries}

\textbf{Causal Model (CM)} formalized by Pearl~\cite{pearl2009causality}, consists of a Bayesian network or directed acyclic graph (DAG), where nodes denote observed variables (covariates) and edges encode causal relationships, quantified by conditional probabilities or expectations. Discovering causal structure requires determining the presence and directions of edges by matching conditional independence relations in the data, guided by the key assumptions of \emph{causal sufficiency}\ie~all relevant variables, especially common causes, are observed, so there are no hidden confounders, \emph{faithfulness}\ie~the observed independencies are due solely to the structure of the causal graph, not accidental parameter cancellations, and the \emph{causal Markov condition}\ie~each variable is independent of its non-descendants given its immediate parents in the graph. Together, these principles enable rigorous identification and analysis of causal effects within a modeled system. Finally, setting nodes to desired values, i.e., \textit{interventions}, allows estimating the weight of edges as causal effects.\\
\textbf{Reinforcement Learning (RL)} enables agents to learn optimal behavior by taking action within an environment, while aiming to maximize cumulative reward over episodes~\cite{sutton2018reinforcement}. Formally, RL problems consist of a Markov Decision Process (MDP), defined by the tuple $(S,A,P,R,\gamma)$, where $S$ is the set of states, $A$ is the set of actions, $P(s'|s,a)$ is the transition probability, $R(s,a)$ is the reward function, and $\gamma \in [0,1]$ is the discount factor; the agent learns a policy $\pi(a|s)$ specifying the probability of taking action $a$ in state $s$.

\textbf{Multi-Agent Reinforcement Learning (MARL)} consists of agents learning within a shared environment~\cite{zhang2021multi} and each seeking to maximize its own cumulative reward. Agents face challenges, including non-stationarity,~i.e., agents alter the environment~\cite{bloembergen2015evolutionary}, scalability issues,~i.e., the joint state-action space grows exponentially with the addition of new agents, and partial observability, i.e., limited access to a global state. As a mitigation, MARL agents can still cooperate by sharing knowledge about their overlapping state spaces~\cite{cooperative_multi_agent_transfer_2021}. \textbf{Heterogeneous Goals (HG)} refer to scenarios in multi-agent systems where agents pursue distinct, non-adversarial objectives that may partially overlap or differ in priorities, roles, or start and end conditions. This diversity requires agents to adapt their strategies and knowledge transfer mechanisms to accommodate varying individual goals within shared environments.  

 \textbf{World Representation (WR)}\cite{russell2010artificial} is an explicit internal model that agents use to predict and reason about state transitions, actions, and outcomes within their environment, often constructed as a causal or state-transition model. Unlike model-free approaches, this representation enables agents to anticipate the effects of their actions and supports informed planning and decision-making. \textbf{Discrete-tabular Environment (DTE)}  is a type of world representation that consists of a finite set of well-defined states, actions, and transitions—such as a grid-world or finite MDP—where all possible interactions can be explicitly enumerated and analyzed. This structure simplifies the modeling, evaluation, and development of reinforcement learning algorithms by providing a manageable and interpretable state-action space. 

\section{The State-of-the-Art}\label{sec:state-of-the-art}

Although recent methods support knowledge acquisition and transfer for safe agent behavior, limitations remain due to the distributed and autonomous nature of agents in non-stationary environments. To highlight this gap, we analyzed and classified existing approaches to transfer learning in loosely coupled multi-agent systems according to key criteria introduced in \Cref{sec:preliminaries}, revealing significant variations in how current methods address these challenges (see \Cref{table:sota}).
 \begin{table}[]
           \centering
            \caption{Synoptic Table Related Work. World Represent. (WR), Discrete/Tabular Env. (DTE), Heterogeneous Goals (HG)}
            \label{table:sota}
            \vspace{-2mm}
            \resizebox{0.99\columnwidth}{!}{
                    \begin{tabular}{lcccc}\hline
                         Approach & WR & DTE & HG \\\hline
                                 MAPTF \cite{efficient_transfer_learning_2021} & $\bullet$ & $\bullet$ & $\bullet$  \\ 
                                 \hline
                                 MALT \cite{lateral_transfer_learning_2021}, Co-MACTRL \cite{co_evolving_multi_agent_rl_2024}, CTL \cite{scenario_independent_representation_2023} & $\bullet$ &  & $\bullet$\\ 
                                   \hline
                                    UPDeT \cite{updet_2021} ,  MACKRL \cite{multi_agent_common_knowledge_rl_2019}& $\bullet$&  & \\ 
                                 \hline
                               MT-SAC \cite{multi_agent_transfer_rl_traffic_2021}, PPO-MLP \cite{evaluating_generalization_transfer_marl_2021}, DDQN \cite{distributed_rl_uav_swarm_control_2020},  FCQN \cite{tensor_action_spaces_2020}   &  & $\bullet$ & $\bullet$ \\ 
                                 \hline
                                 3D CNN MARL \cite{multi_agent_pathfinding_3D_2024}, MADDPG-SFKT \cite{efficient_exploration_successor_features_2022} &  & $\bullet$ &  \\ 
                                   \hline
                                   TL in MOD-RS\cite{multi_agent_ride_sharing_systems_2021} &  &  & $\bullet$ \\ 
                                 \hline
                         \textbf{Our Approach} & \textbf{$\bullet$} & \textbf{$\bullet$} & \textbf{$\bullet$}\\\hline
                    \end{tabular}
                    }
                    \vspace{-5mm}
        \end{table}


The literature review suggests that six papers \cite{efficient_transfer_learning_2021, multi_agent_common_knowledge_rl_2019, updet_2021,lateral_transfer_learning_2021,co_evolving_multi_agent_rl_2024, scenario_independent_representation_2023} utilize an explicit world representation (WR) to model the environment and inform decision-making. 
MAPTF\cite{efficient_transfer_learning_2021} models the knowledge
transfer among agents as the \emph{option learning problem} to determine which agent’s policy is useful for each agent, and when to terminate it to avoid \emph{negative transfer}. Option learning refers to learning temporally extended actions\ie~options, rather than primitive, one-step actions. Negative Transfer happens when, after knowledge transfer, the new knowledge from one task negatively impacts the performance of another task \cite{zhang2022survey}. However, while MAPTF supports heterogeneous goals (HG), it shares one option set across agents, assuming agents are homogeneous and their policies are equally reusable—a constraint in heterogeneous teams. Additionally, the framework focuses on local agent perspectives; it doesn’t address global coordination or joint credit assignment, which is crucial in many MARL settings. 
MALT~\cite{lateral_transfer_learning_2021} proposes a lateral transfer mechanism for enhancing cooperation, relying on a world representation to transfer knowledge efficiently between agents. Co-MACTRL~\cite{co_evolving_multi_agent_rl_2024} utilizes a scenario-independent representation for co-evolving systems, making it possible to apply knowledge across evolving scenarios. CTL~\cite{scenario_independent_representation_2023} uses a state- and action-independent representation, enabling flexible knowledge transfer across tasks, supporting both homogeneous and heterogeneous agents; however, due to its use of deep neural networks, it does not support tabular environments that expect finite, countable state-action pairs. 

UPDeT~\cite{updet_2021} introduces a transformer-based model for policy decoupling, leveraging a world representation to predict and adapt policies in dynamic environments. The approach focuses on individual policy learning, achieving generalizability, and explainable learning. However, it does not account for personalized objectives or agent-specific rewards and coordination, and full heterogeneity is not supported. 
In MACKRL \cite{multi_agent_common_knowledge_rl_2019}, decentralized agents coordinate based on a shared common knowledge. However, these agents cooperate with shared goals, instead of individual, heterogeneous ones. 

Four of the surveyed approaches operate in discrete or tabular environments, but do not support an explicit WR. For example, the traffic control optimization framework by \cite{multi_agent_transfer_rl_traffic_2021} uses a multi-view encoder for adaptive traffic signal control in a structured, discrete setting. Similarly, PPO-MLP~\cite{evaluating_generalization_transfer_marl_2021} evaluates the generalization of transfer learning across various numbers of agents in a grid-like environment. These environments, often based on grid-world setups or finite MDPs, allow for easier analysis and structured testing of algorithms. Works such as \cite{distributed_rl_uav_swarm_control_2020} and \cite{tensor_action_spaces_2020} also adopt discrete environments to explore multi-agent learning; the approach in~\cite{multi_agent_pathfinding_3D_2024} even uses a discrete 3D environment for path finding.

Another important area of distinction among these approaches is how they handle heterogeneous goals (HG) in multi-agent systems. For example, in traffic control, \cite{multi_agent_transfer_rl_traffic_2021}, agents have different but coordinated goals in dynamic environments. Similarly, \cite{lateral_transfer_learning_2021} emphasizes cooperation among agents with diverse roles. In more recent works, such as \cite{co_evolving_multi_agent_rl_2024}, the focus is on leveraging knowledge transfer across agents with evolving goals and scenarios, demonstrating the importance of supporting heterogeneous objectives. The research by \cite{multi_agent_ride_sharing_systems_2021} on ride-sharing systems further exemplifies how agents with different goals can still benefit from shared knowledge to enhance performance. 
 MADDPG-SFKT~\cite{efficient_exploration_successor_features_2022}  offers knowledge transfer across tasks, however, similar to UPDeT~\cite{updet_2021}, MADDPG-SFKT does not support HG, and the performance of the knowledge transfer depends on how closely target and source tasks align during training. 

Given the observations in~\Cref{table:sota}, there remains no method that (1) operates in a discrete/tabular grid environment, (2) supports agents with heterogeneous, non‐adversarial goals, and (3) builds an explicit, interpretable world representation, via causal discovery for offline macro‐action selection and transfer. 
While MAPTF~\cite{efficient_transfer_learning_2021} satisfies the three criteria in~\Cref{table:sota}, it emphasizes expressive, flexible policy representations through neural networks, which are less interpretable and require learning on each new task, albeit more sample-efficient than from scratch. In this paper, we put forward a lightweight, interpretable, and reusable causal structure. 
Finally, this work aims to fill this gap by (1) treating each collision as a small causal experiment, (2) estimating causal effects of each recovery macro, and (3) allowing agents to reuse these macros without further learning. This combination of discrete interpretability, heterogeneous goals, and rigorous causal modeling is, to our best knowledge, novel in the MARL literature.

\section{Approach} \label{sec:approach}

The proposed causal transfer learning framework for MARL follows three insights: (1) leverage on the shared partially observable state-space of obstacle regions, (2) abstract trajectories as causal pathways, and (3) consider adaptations as recovery actions. These insights are realized in a hierarchical architecture to support three main function (stages)\ie~\emph{represent}, \emph{acquire}, and \emph{transfer} knowledge among agents.\footnote{The implementation and experimental setup are publicly available at: \href{https://github.com/kat-ko/Causal-Based-Knowledge-Transfer-for-Multi-Agent-Reinforcement-Learning-in-Dynamic-Environments}{github.com/kat-ko/CausalMARL}.}


The three stages constitute the main building blocks of the proposed causal transfer learning framework and feed directly into one another. First, during \emph{recovery–action discovery} \Cref{subsec:acquisition} each agent's collision is logged together with the sequence of movements that resolves it, yielding a dataset of $(\text{context},\text{action},\text{outcome})$ triples. Second, the \emph{causal–effect estimator} \Cref{subsec:representation}) processes this dataset offline after: for every collision context, we can select the macro that minimizes the expected residual path length and write the result into a compact lookup table. Third, in the \emph{transfer stage} (\Cref{subsec:transfer}), any agent can consult this table at run time; the retrieved macro is executed as an atomic block, and control then returns to the agent’s baseline policy. Taken together, these stages convert raw interaction into portable recovery knowledge.

\subsection{Representation of Recovery Action Knowledge}
\label{subsec:representation}
\textit{Recovery Action (RA) macros} are sequences of actions that allow an agent to overcome obstacles after a collision $\chi_i$. In our framework, each macro is recorded from the point following a collision until reaching the next collision or the agent's goal, and is associated with the local context in which it was applied. Knowledge transfer occurs when an agent selects and executes an RA macro stored in a \textit{lookup model}, which indexes the local context of a collision $\chi_i$, e.g., state and downstream of recovery actions that minimize the expected remaining path length towards various goals. To mitigate the cold start problem, i.e., insufficient transfer knowledge, we pre-trained a collection of agents in the same environment but pursuing heterogeneous goals. To compute the impact of RA macros on an agent's cumulative reward, we represented the environment paths $\Psi$ via adopting the graph formalisms of \textit {Causal Models (CM)}---see~\Cref{sec:preliminaries}. Each version of the environment is instantiated as a $CM_i$. The application of an RA macro consists of an \emph{intervention} (see~\Cref{sec:preliminaries}) on a $CM_i$. Hence, each obstacle collision $\chi_i$ can be understood as a localized causal experiment, where a causal path $\Psi_i$ is selected from the $CM_i$ to encode the relationship between the collision context (state and direction), the chosen sequence of recovery actions, and the final outcome\ie~total path length to the goal. The \textit{Causal Effect Estimator (CE)} approximates the reward gain of applying a given RA macro, ensuring that the ranking of actions reflects their actual impact on the agent’s efficiency, including all downstream effects after the intervention.

\noindent\textbf{Formalization of Recovery Action Knowledge} -- To systematically model and evaluate the impact of obstacles on agent behavior, we introduce a modular abstraction of agent trajectories based on collision events. Let \( k \in \mathbb{N}_0 \) denote the number of collisions an agent experiences during an episode. In an unperturbed environment without obstacles, the agent $i$ follows its optimal path $\Psi^*_{i}$ from start $\alpha_i$ to end $\Omega_i$ without any collisions, corresponding to \( k = 0 \). The path structure in this case can be abstracted as $\alpha \rightarrow \Psi^*_{i} \rightarrow \Omega$. When obstacles are introduced, each collision $\chi_i$ divides the trajectory to pre- and post-collision paths, respectively $\Psi^{\rho}_{i}$ and $\Psi^{\tau}_{i}$. Hence, the path is now $\alpha \rightarrow \Psi^{\rho}_{i} \rightarrow (\chi_i \rightarrow \Psi^{\tau}_{i})^{k}_{i=1} \rightarrow \Psi^{\rho}_{k+1}  \rightarrow  \Omega$.  
            
The cumulative episode reward, given that every valid movement incurs a \(-1\) penalty and the goal yields \(0\) reward, reflects the total path length. The relation between rewards before and post-collision events can thus be expressed as $R = \rho + \sum_{i=1}^{k} \tau_i$, where $R$ is the total path reward, $\rho$ is the pre-collision reward, and $\tau$ is the post-collision reward. Given that RA macros may vary in efficiency, the total episode length (and thus the final reward) depends critically on the quality of the chosen recovery macros. 
            
In the limiting case where no obstacles are present (\(k=0\)), the agent simply follows the globally optimal path $\Psi^*_{i}$. In scenarios where \(k \geq 1\), the agent must adapt by selecting RA macros at each collision $\chi_i$, thus turning the overall planning into a sequence of local causal intervention subproblems. This modularization naturally lends itself to a divide-and-conquer approach. By solving and optimizing each recovery subproblem independently, the agent can reconstruct near-optimal behavior in complex, perturbed environments. 
  
\subsection{Acquisition of Recovery Action Knowledge}
\label{subsec:acquisition}
\noindent\textbf{Offline Exploration} -- Agents are first pre-trained in an environment without obstacles to learn baseline policies.\\
\noindent\textbf{Recovery Action Discovery} -- When obstacles are introduced, agents switch to exploratory behavior upon each collision, recording the sequences of actions\ie~recovery actions,  taken to bypass the obstacle. Each recovery action is logged along with its context and the resulting path length (see~\Cref{algo:ra-discovery}).

\noindent\textbf{Causal Model Fitting} -- The agents' experience\ie~actions taken within an environment, is used to fit a causal model\footnote{We adopted a doubly-robust estimators CausalForestDML \cite{athey2019estimating} with $n_{estimators}=200$, $min_{samples.leaf}=10$, $\alpha=1.0$, $cv=3$ and $max_{iter}=500.$} while adjusting for confounders (e.g., collision state, attempted movement direction, prior path length) to estimate the expected outcome for each RA macro, i.e., a causal knowledge \textit{CK} within a local context (a collision $\chi_i$).\\
\noindent\textbf{Formalization of Causal Knowledge} -- Because of the interplay between different start and end position coordinates and obstacles, agents' minimal paths differ. To capture the performance of path-finding between any given path $L_{agent}$ and the optimum $L_{opt}$, we proposed a metric called optimal-to-final path-length ratio \textit{OFPR} $= \frac{L_{\text{opt}}}{L_{\textit{agent}}}$. An OFPR value of 1.0 implies the agent found an optimal path, whereas lower values indicate how \textit{inefficient} its route was (e.g., 0.8 means it took 25\% more steps than optimal).

\subsection{Transfer of Recovery Action Knowledge}
\label{subsec:transfer}
\noindent\textbf{Teacher-Learner Transfer} -- The recovery action lookup model learned by a teacher agent is shared with a learner agent operating 
within the same obstacle environment. The learner can receive causal knowledge from multiple teachers, and a teacher can provide knowledge to multiple learners. This model of transfer extends beyond the simpler dyadic transfer relations (source-target) to many-to-many ones.

\noindent\textbf{Zero-Shot Application} -- Upon encountering a collision, the learner agent queries the transferred lookup model using its current collision context and executes the recommended recovery action as a macro-action, without any additional online learning or adaptation.\\
\noindent\textbf{Isolation of Transfer} -- To ensure that performance improvements stem only from transfer rather than joint adaptation, the teacher model is not updated with the learner experiences.\\
\noindent\textbf{Metric for Causal Knowledge Transfer} -- We compute a delta metric $\Delta_{\text{CK}} = T_{CK}-L_{CK}$ to denote the \textit{net teacher value} that isolates complementary aspects of transfer effectiveness. Let $L_{CK}$ be the OFPR of the learner (the agent who receives the causal knowledge) and $T_{CK}$ the OFPR of the learner after importing the teacher’s causal model. A value of 0 then implies no difference between $T_{CK}$ and $L_{CK}$, whereas positive values denote a transfer benefit, while negative values denote a detrimental effect.
    
\small
\begin{algorithm}[t]
\caption{Recovery-Action Discovery}\label{algo:ra-discovery}
\begin{algorithmic}[1]
\STATE \textbf{Input:} obstacle scenarios $\mathcal{S}$,\; pre-trained tables $Q_{\pi}$ 
\FOR{each scenario $s \in \mathcal{S}$}
    \STATE initialize RecoveryAgent$(Q_{\pi})$
    \FOR{episode $1 \ldots N_{\text{RA}}$}
        \STATE reset environment;\; $in\_RA \gets \textbf{false}$
        \WHILE{episode not finished}
            \STATE $a \gets$ agent.act($state$)
            \STATE $(state',r,collide)\gets$ env.step($a$)
            \IF{$collide$ \AND \textbf{not} $in\_RA$}
                \STATE start RA record;\; $in\_RA \gets \textbf{true}$
            \ELSIF{$in\_RA$ \AND \textbf{not} $collide$}
                \STATE append $a$ to current RA sequence
            \ELSIF{$in\_RA$ \AND $collide$}             
                \STATE save current RA;\; start next RA record
            \ENDIF
            \STATE $state \gets state'$
        \ENDWHILE
    \ENDFOR
\ENDFOR
\end{algorithmic}
\end{algorithm}
\normalsize

\section{Evaluation}\label{sec:evaluation}

\subsection{Experimental Setup}

\noindent\textbf{Goals} -- When perturbations modify state spaces shared by multiple agents, the corresponding changes in agents’ policies can provide insights into how generalized causal knowledge can be applied. To investigate how differences in goals affect transferability, we define four canonical scenarios: \textit{SS-SE} (Same Start \& Same End), 
\textit{SS-DE} (Same Start \& Different End), 
\textit{DS-SE} (Different Start \& Same End),  
and \textit{DS-DE} (Different Start \& Different End). 

\begin{figure}[t]
\vspace{-3mm}
    \centering
    \includegraphics[width=\linewidth]{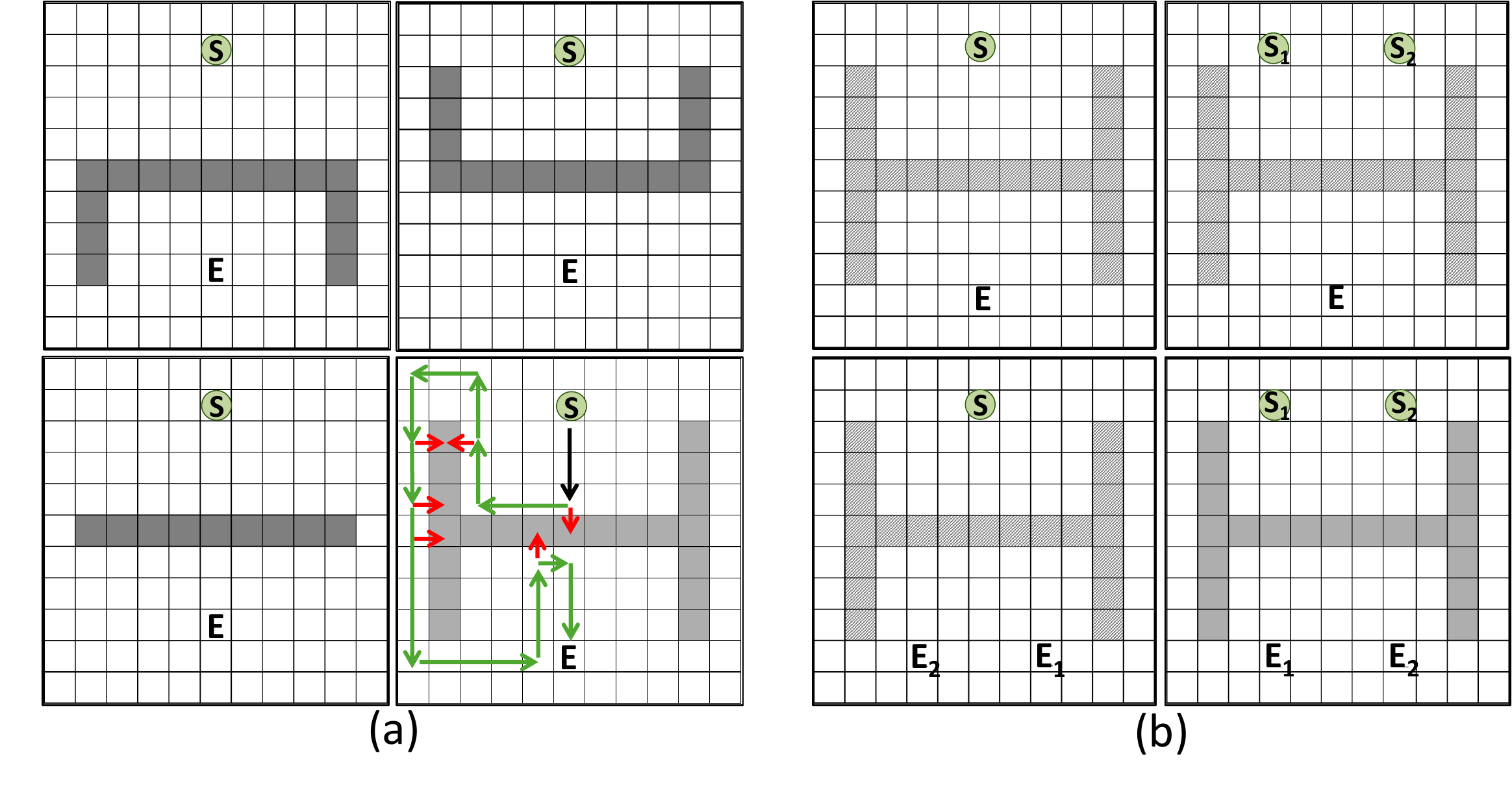}
    \caption{Schematic visualization of the 11 x 11 grid-world. (Sub-figures are described from top left to bottom right. (a) Obstacle scenarios -- Reverse U, U, Wall, Superposition. Bottom right shows an example of an agent's movements during recovery action discovery during training a policy $\pi_{CK}$. The agent follows its pre-trained policy (black arrow) until collision, then explores the grid, logging macro-actions (green arrows) and attempted movements leading to collisions (red arrows). (b) Goal scenarios -- agent start-end coordinates across the four scenarios: SS-SE, DS-SE, SS-DE, DS-DE.}
    \label{fig:gridworld}
    \vspace{-5mm}
\end{figure}

\noindent\textbf{Barriers} -- A curriculum of obstacle configurations allows varying the difficulty of goals (\Cref{fig:gridworld}-(a)), i.e., introducing a progressively greater challenge to agent navigation. The \emph{Wall} obstacle represents a simple planar barrier that can be circumvented with minimal detour. The \emph{Reverse U} presents a vertical extensions that partially enclose the path. The \emph{U} obstacle further amplifies difficulty by forcing agents into a constrained corridor, demanding multi-step detours to escape. 

\noindent\textbf{Baselines} -- For every barrier scenario, three reference agents are compared systematically within and across goal scenarios to establish how effectively causal recovery knowledge can augment agents. (1)~\textit{Rand} (random exploration, lower bound): agent with no prior knowledge (pre-training or recovery macros) randomly explores upon encountering barriers. 
(2)~$\pi_{CK}$ (causal model augmented agent): agent pre-trained with Q-Learning without barriers and equipped with the causal model (CM) built offline. Upon collision, it queries the lookup model for the best recovery action. (3)~$P^*$ (full retraining, upper bound): agent trained with Q-Learning with barriers, representing the optimal attainable policy within barrier environments.

\noindent\textbf{Scenario} -- We investigate how the causal recovery knowledge is transferable across agents by subjecting teacher and learner agents to a permutation of barriers and goals, see~\Cref{fig:gridworld}-(a), and~\Cref{fig:gridworld}-(b). We fit a causal model for a teacher agent in a given goal, then we transfer it to a learner agent operating under a different goal, while sharing the same barrier layout. The learner agent uses its pre-trained obstacle-free policy ($Q_{\pi}$) and relies on the teacher’s RA macro to look up upon collision experiences. Hence, the resulting agent is the learner $\pi^{L}_{CK}$ equipped with the teacher causal model $\pi^{T}_{CK}$.

\subsection{Answers to Research Questions}\label{sec:results}

\begin{figure}[t]
\vspace{-3mm}
    \centering
    \includegraphics[width=0.85\linewidth]{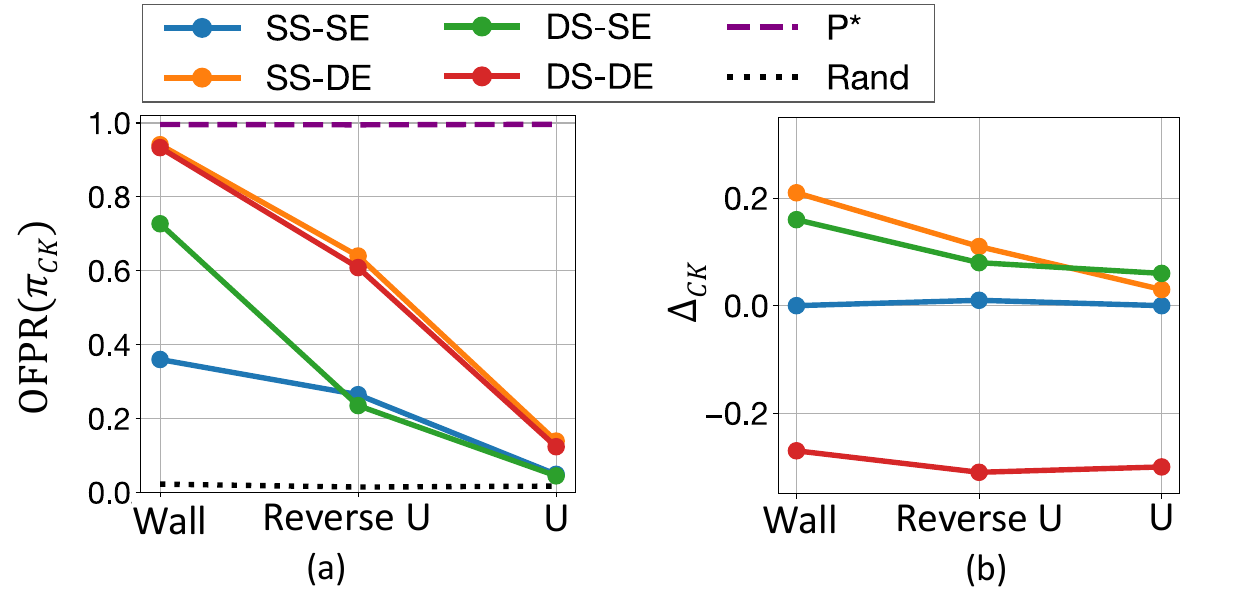}
    \caption{The X-axis represents the barrier difficulty. (a) Shows the path efficiency when using the acquired causal knowledge (OFPR metric of policy $\pi_{CK}$ plotted against the $P^*$ and Rand baselines) for each start-end configuration across barrier scenarios. (b) Net teacher value ($\Delta_{CK}$ metric) after knowledge transfer for each start-end configuration across barrier types.}
    \label{fig:OFPR-Transfer}
    \vspace{-5mm}
\end{figure}

\textbf{RQ1}\textit{ -- Does environment complexity impact the efficiency of causal-based knowledge transfer?} \textbf{Yes}, agents subject to increasingly complex barriers show a downward trend of the OFPR($\pi_{CK})$, see~\Cref{fig:OFPR-Transfer}-(a) where increasing difficulty also lowers the OFPR variance across goal scenarios, caused by geometry-induced optimal path-length offsets.

\textbf{RQ2}\textit{ -- Does heterogeneity of goals impact the efficiency of causal-based knowledge transfer?} \textbf{Partially yes}, as shown in \Cref{fig:OFPR-Transfer}-(a), heterogeneity of goals mostly impacts OFPR($\pi_{CK}$) for agents transferring causal knowledge when subject to simpler barriers\eg~Wall. However, this behavior is also not uniform across goal scenarios. Note that same ends (SE) seems to pose more challenges for benefiting from causal transfer than having different ends (DE). Intuitively, heterogeneity of goals interacts with barrier complexity and in ways that our causal modeling could not capture. 


\textbf{RQ3} \textit{ -- Is the difference in causal-based knowledge transfer sensitive to environment complexity?} \textbf{Partially yes}, for the two goal scenarios in which transfer is beneficial, as shown in \Cref{fig:OFPR-Transfer}-(b), there is a downward trend in $\Delta_{CK}$ from simpler to more complex environments. For both the least and most heterogeneous goal scenarios (blue and red lines, respectively), the causal knowledge gap seems to be insensitive to environment complexity.

\textbf{RQ4} \textit{ -- Is the difference in causal-based knowledge transfer sensitive to goal heterogeneity?} \textbf{Yes}, as shown in \Cref{fig:OFPR-Transfer}-(b). In Alignment with the expected potential for benefiting from the teachers acquired knowledge for the learner dependent on goal similarity, (1) the least heterogeneous goal (blue line) shows no transfer benefit $\Delta_{CK}=0$, while (2) the most heterogeneous goal (red line) shows a detrimental impact of causal transfer $\Delta_{CK}<0$.\\
\noindent\textbf{Take-aways} -- These results point to two key insights: (RQ1, RQ2) agents with heterogeneous goals on average closed about half the gap between random exploration and a fully retrained policy when adapting to new environments, and (RQ3, RQ4) the success of causal knowledge transfer depends on the interplay between barrier complexity and goal similarity.

\section{Discussion and Conclusion}\label{sec:discussion-conclusion}

\subsection{Implications}
The causal modeling of knowledge transfer enabled us to systematically examine the effects of environmental changes on agents’ strategies as a means of \textit{learning to transfer through obstacles}. This is a stepping stone to investigate the concept of \textit{causal knowledge transfer} as an instance of \textit{curriculum learning}. The interplay shown between goal heterogeneity and environment complexity could be used to search for new compositions of causal knowledge, in which multiple causal models contribute to a single recovery action macro. The ability to leverage causal knowledge from multiple sources can further support a more \textit{light-weight implicit coordination} model, where agents cooperate towards a common goal or fight back adversarial agents that mislead each other by adding obstacles or taking confounding recovery actions.

\subsection{Threats to Validity} Generalizability (external threat to validity) was limited to differences in goals and environments. We plan to extend this to agents with distinct skill sets and more diverse types of barriers, e.g., soft ones and cliffs. Internal validity was mitigated by preventing confounding of causal effects. However, we assumed causal sufficiency (or ignorability, i.e., no hidden confounders). Future work could explore relaxations of this assumption and evaluate its impact (sensitivity analysis) on knowledge transfer efficiency. Finally, we mitigated the risk of the wrong object of study measurement (construct validity) by posing different metrics (OFPR and $\Delta_{CK}$). Novel metrics are needed for the failure modes of knowledge transfer, e.g., catastrophic forgetting and negative transfer.

\subsection{Conclusion and Future Work}\label{sec:conclusion}

We addressed the challenge of knowledge transfer in non-stationary multi-agent reinforcement learning (MARL) environments, where agents must adapt to varying goals and obstacle configurations. Traditional policy- or value-based transfer methods often fail to generalize in such settings due to their lack of explanatory structure. We proposed a causal transfer learning framework that treats agent-obstacle interactions as structural causal experiments, enabling the offline estimation and transfer via recovery action macros, i.e., causal interventions that minimize detour costs. By indexing recovery actions in a lookup model and enabling zero-shot application of these actions, we enabled agents to reuse knowledge across heterogeneous goal scenarios without retraining. 
These contributions advance the state of causal MARL by offering an interpretable, efficient, and decentralized mechanism for safer knowledge transfer. Future work will extend the causal modeling framework to support more diverse obstacle types, relax the causal sufficiency assumption via sensitivity analysis, and explore curriculum learning strategies to prioritize transfer sources based on structural path similarity.


\bibliography{references}


\bibliographystyle{abbrv}


\end{document}